# New wrapper method based on normalized mutual information for dimension reduction and classification of hyperspectral images

Hasna Nhaila*
Electrical Engineering Research Laboratory
ENSET. Mohammed V University
Rabat, Morocco
hasnaa.nhaila@gmail.com

Asma Elmaizi
Electrical Engineering Research Laboratory
ENSET. Mohammed V University
Rabat, Morocco
asma.elmaizi@gmail.com

Elkebir Sarhrouni
Electrical Engineering Research Laboratory
ENSET. Mohammed V University
Rabat, Morocco
sarhrouni436@yahoo.fr

Ahmed Hammouch
Electrical Engineering Research Laboratory
ENSET. Mohammed V University
Rabat, Morocco
hammouch_a@yahoo.com

*Abstract*—Feature selection is one of the most important problems in hyperspectral images classification. It consists to choose the most informative bands from the entire set of input datasets and discard the noisy, redundant and irrelevant ones. In this context, we propose a new wrapper method based on normalized mutual information (NMI) and error probability (PE) using support vector machine (SVM) to reduce the dimensionality of the used hyperspectral images and increase the classification efficiency. The experiments have been performed on two challenging hyperspectral benchmarks datasets captured by the NASA's Airborne Visible/Infrared Imaging Spectrometer Sensor (AVIRIS). Several metrics had been calculated to evaluate the performance of the proposed algorithm. The obtained results prove that our method can increase the classification performance and provide an accurate thematic map in comparison with other reproduced algorithms. This method may be improved for more classification efficiency.

*Keywords— Feature selection, hyperspectral images, classification, wrapper, normalized mutual information, support vector machine.*

## I. INTRODUCTION

With the recent development on hyperspectral sensors technologies, the hyperspectral images (HSI) become more available and widely employed in many applications in different domains such as food industry [1], military [2], agriculture mapping and especially land cover analysis [3].

In real world applications, hyperspectral images are represented by more than a hundred of bands of the same observed region. In the classification schemes, this large amount of spectral information increases the discrimination between classes. Unfortunately, it possesses many challenges in treatment and processing time due to the presence of redundant, irrelevant bands and the limited number of training samples. This problem is known as curse of dimensionality [4]. Subsequently, the dimension reduction (DR) becomes a crucial preprocessing step of HSI classification. The DR may be done either by selection, extraction or selection followed by extraction [5]. Several works had been done in this context [2], [6]. In this study, we use feature selection methods that can be divided into two main categories: filter and wrapper according to the relationship between the feature selection method and the induction algorithm (classifier). In this study we use the wrapper approach.

The rest of this article is organized as follows. The next section presents the related works of HSI dimensionality reduction using feature selection. In section 3, we explain the proposed algorithm based on both normalized mutual information NMI and error probability PE with SVM classifier. The used datasets and discussion about experiments are presented in section 4. Finally, section 5 concludes our work.

## II. RELATED WORK

Dimensionality reduction using feature selection is an important step in hyperspectral images classification. It consists to reduce the complexity of input data by selecting the main informative features. Among the feature selection methods presented in the literature, mutual information (MI) based algorithms are the most popular in many HSI applications. For this, several approaches have been proposed. In [7], maximum relevance minimum redundancy algorithm (MRMR) was proposed to select good features according to the maximal statistical dependency criterion based on mutual information. Guo in [8] used MI to select bands for hyperspectral image fusion. And in his work [9], he proposed a fast feature selection scheme based on a greedy optimization strategy. Additionally, in [10], a novel unsupervised clustering is applied on HSI based on the similarity measure and histogram. Following this works, in [11], a wrapper algorithm using mutual information and inequality of Fano was proposed. In [12], combined mutual information with homogeneity feature extracted from Grey Level Co-occurrence Matrix (MIH) was used to select features from the HSI. New algorithms are constantly appearing; In [13], a hierarchical band selection approach by constructing a spectral partition tree based on mutual information was proposed. In our work, we propose a new wrapper algorithm (WNMIPE) based on normalized mutual information NMI and error probability PE with support vector machine SVM to reduce the dimensionality of the used hyperspectral datasets to increase the classification efficiency. Performance evaluation of the proposed method is performed on two hyperspectral datasets: Indian Pines and Salinas provided by the NASA's AVIRIS sensor. The proposed algorithm is compared with three other reproduced feature selection methods.

The novelty of our contribution is the use of the normalized mutual information in a wrapper approach for hyperspectral images dimensionality reduction and classification using support vector machines. The second novelty is the use of error probability with NMI as evaluation criteria for redundancy control of selected bands.

III. METHODOLOGY

This work proposes a new wrapper methodology to select the most informative bands from the used hyperspectral datasets. It is based on three steps:
- Computing the normalized mutual information NMI as described in the next section.
- Using the wrapper approach with SVM as induction algorithm to construct the reduced subset of bands.
- Applying the error probability PE as an evaluation criterion to improve the classification performance.

A. *Normalized mutual information*

In the context of hyperspectral images, the mutual information is the statistical measure of similarity between the reference (ground truth in our case) noted G and each band noted B.

The mutual information between G and B is given as:

$$I(G,B) = \sum log_2 p(G,B) \frac{p(G,B)}{p(G).p(B)} \quad (1)$$

In relation with Shannon entropy, the MI gives a measure of dependence by calculating the difference between the independent and joint distributions of the entropy as defined in equation 2:

$$MI(G,B) = H(G) + H(B) - H(G,B) \quad (2)$$

If the ground truth and the band are independent (the case of noisy bands for example), their joint distribution is equal to the sum of their individual distribution. The MI takes values from zero (independent variables) and $+\infty$ (largest information shared between the variables). When applying on hyperspectral images, the MI can be limited by the total amount of information in images and becomes hard to interpret due to the unbounded range of values. To solve this, we use one of the various measures of the normalized mutual information NMI defined as the ratio of the entropy of G and B on the joint entropy between G and B as in equation 3:

$$NMI(G,B) = \frac{H(G)+H(B)}{H(G,B)} \quad (3)$$

The required probabilities for $p(G)$, $p(B)$ and $p(G,B)$ are estimated using a histogram of the intensity distribution values. The normalization of mutual information scales values of mutual information in a bounded range [0, 1] with:
- Value 1 means a perfect correlation between the ground truth map and the band.
- Low values indicate a small similarity.
- Zero shows that the two variables G and B are independent.

B. *Error probability*

To control the redundancy in our wrapper scheme, we use the error probability PE proposed in [14] [11] and expressed as follows:

$$\frac{H(G\backslash B)}{log_2 N_c} \leq PE \leq \frac{H(G\backslash B)}{log_2} \quad (4)$$

We compute the normalized mutual information between the ground truth G and the subset of candidate bands B with $N_c$ is the number of classes in the used dataset. This concept can be formulated as:

$$PE \leq \frac{H(G)-I(G,B)-1}{log_2 N_c} = \frac{H(G\backslash B)-1}{log_2 N_c} \quad (5)$$

In our selection process, since $N_c$ and $H(G)$ are constant, when $I(G,B)$ is maximal then PE becomes minimal. So the candidate band that minimizes the error probability has a high similarity with the ground truth and minimum redundancy with the already selected bands and it will be added at the selected subset if not it will be discarded.

C. *Proposed algorithm*

Our main idea in this proposed method is that the candidate band B is a good approximation of the ground truth G if it had a higher value of normalized mutual information. Since our algorithm is a wrapper scheme, this band must decrease the computed $PE$ to increase the classification accuracy. So our algorithm is seen as an incremental wrapper based subset selection (IWSS) [15] using normalized mutual information.

Noting that the induction algorithm used for the classification is the support vector machine SVM with RBF kernel. It is one of the most useful as supervised classifier which provided good results in many hyperspectral images classification works [16][17].

The complete selection process of our proposed method is as follows:

**Algorithm**

*Input: hyperspectral dataset*
  *G: Ground truth,*
  *B: dataset Bands*
  *T: Training samples*
*Output:*
  *S: Selected set of bands.*
*1 (Initialisation)*
  $B \leftarrow initial\ set\ of\ input\ features\ (bands)$
  $S \leftarrow Group\ of\ selected\ bands\ "empty\ set"$
  $l \leftarrow The\ number\ of\ bands\ to\ be\ selected$
  $Th \leftarrow Threshold\ to\ control\ redundancy$
*2 (Computation of normalized mutual information MI between G and each band B) using equation 3.*
$For\ b: \in B, compute\ NMI(G, b_i)$
*3 (Selection process)*
*Select the first band $b_i$ that maximizes $NMI(G, b_i)$*
$b_i = argmax_{b_i} NMI(i)$
$Set\ B \leftarrow B \setminus \{b_i\}\ ;\ S \leftarrow \{b_i\}\ ;\ G_{-est0} = band(b_i)$
$Compute\ PE^*\ :\ PE \leftarrow PE^*\ ;$
$while\ [S] < l\ do$
  $b_i = argmax_{b_i \in (B-S)} NMI(i)$
  $Set\quad G_{-est} = G_{-est}(S)$
$Compute\ PE$
$if\ PE \leq PE^* - Th\ \ then\ PE \leftarrow PE^*\ and\ S \leftarrow S \cup \{b_i\}\ ;$
$else\ S \leftarrow S/\{b_i\}\ ;$
$end\ if$
$end\ while$
*5 (Output) S is the set of selected bands.*

## IV. EXPERIMENTS

### A. Datasets description

In this paper, two challenging hyperspectral benchmark datasets are used for experiments from Airborne Visible/Infrared Imaging Spectrometer Sensor (AVIRIS). These datasets have been previously used in other researches [18-19] and are publicly available at http://www.ehu.eus/ccwintco/index.php/Hyperspectral_Remote_Sensing_Scenes. They have different characteristics in terms of dimensions and features type.

*1) Indian Pines dataset*

The first dataset used in this study is acquired over the Indian Pines in North-western Indiana. It has 145x145 pixels and 224 bands in the wavelength range of 0.4-2.5 μm with spatial resolution of 20 m pixels. The Color composite and the corresponding ground truth reference of this dataset are presented respectively in (a) and (b) in Figure 1. It contains 16 classes which are listed in the same figure.

*2) Salinas dataset*

The second dataset used in this article is Salinas. It is captured over Salinas valley, CA, USA. It consists of 217x512 pixels and 224 spectral reflectance bands in the wavelength range of 0.4 to 2.5 μm. Salinas scene is characterized by high spatial resolution (3.7 m pixels). The Color composite and the corresponding ground truth reference of this dataset are presented respectively in (a) and (b) in Figure 2. It contains also 16 classes which are listed in the same figure.

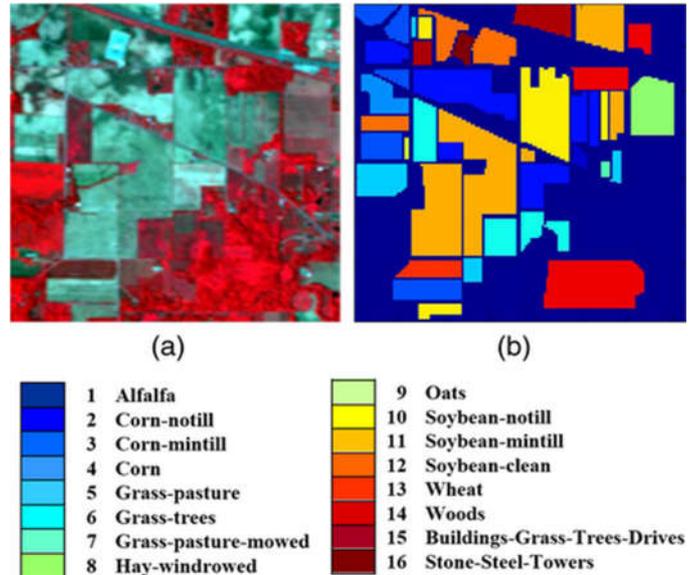

Fig. 1. The Color composite and the corresponding ground truth with class labels for Indian Pines dataset

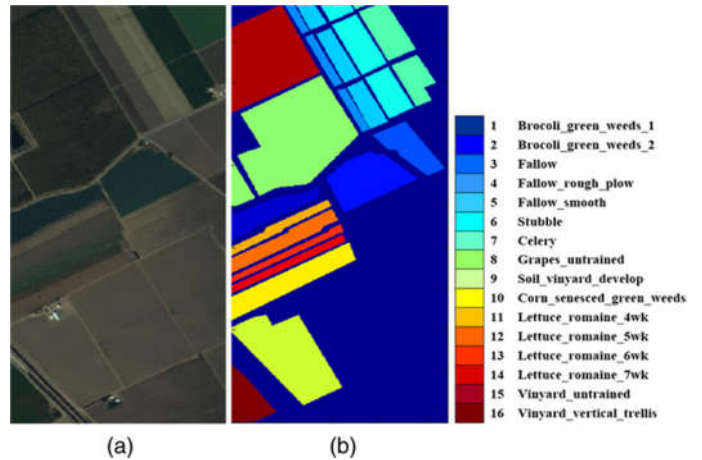

Fig. 2. The Color composite and the corresponding ground truth with class labels for Salinas dataset

### B. Classification and comparison methods

The performance of the proposed algorithm is evaluated in terms of dimension reduction and classification accuracy. In all experiments, we use Support Vector Machine SVM which is a supervised classifier widely used in real world applications of HSI. Radial basis function RBF was chosen as the kernel. The experiments had been compiled in Matlab interface using the Libsvm package to deal with multiclass problems available at www.csie.ntu.edu.tw/cjlin/libsvm. We executed tests on a PC 64-b quad-core Duo CPU 2.1Ghz frequency with 3GB of RAM.

To develop the classification models, the number of samples used for training and testing are made randomly. Three cases have been considered in our study: 10%, 25% and 50% of pixels from each data were used for training and the remaining samples respectively of 90%, 75% and 10% were used to test the models.

In order to validate the obtained results of our proposed method (WNMIPE), we compare it to other two filter approaches including maximum relevance minimum redundancy algorithm (MRMR) and the mutual information with homogeneity tagged (MIH) [12]. Also a wrapper algorithm using mutual information and inequality of fano tagged (WMIF) [11] is considered in the comparison.

*C. Results and discussion*

The experimental results of the proposed approach on Indian Pines and Salinas datasets are presented in this section and are assessed using four evaluation metrics. Individual Class Accuracy (ICA) which represents the correctly classified pixels for each class, Average Accuracy (AA) which is the average of classification accuracy for all classes, Overall Accuracy (OA) which refers to the correctly classified pixels over all test samples and the Kappa coefficient ($k$) of agreement [20].

Table I and II show the classification results in terms of AA, OA and Kappa coefficient obtained by the proposed method for respectively the Indian Pines and Salinas datasets.

TABLE I. AA(%), OA(%) AND KAPPA COEFFICIENT OBTAINED BY THE PROPOSED METHOD FOR THE INDIAN PINES DATASET USING 49 SELECTED BANDS WITH THREE TRAINING SETS.

|       | *10% training* | *25% training* | *50% training* |
|-------|---------------|---------------|---------------|
| AA    | 81.10         | 81.17         | 81.61         |
| OA    | **87.61**     | **88.58**     | **90.29**     |
| Kappa | **0.8679**    | 0.8782        | 0.8964        |

The first step was to select the relevant bands from these datasets that contain 224 bands.

TABLE II. AA(%), OA(%) AND KAPPA COEFFICIENT OBTAINED BY THE PROPOSED METHOD FOR SALINAS DATASET USING 37 SELECTED BANDS WITH THREE TRAINING SETS.

|       | *10% training* | *25% training* | *50% training* |
|-------|---------------|---------------|---------------|
| AA    | 96.46         | 97.44         | 97.90         |
| OA    | **92.55**     | **94.44**     | **95.56**     |
| Kappa | **0.9206**    | 0.9407        | 0.9526        |

From tables I and II, we can make the following remarks:

- Our band selection method provides satisfactory results of classification with overall accuracy that achieves 87.61%, 88.58% and 90.29% for respectively 10%, 25% and 50% as training samples in the Indian pines. For Salinas, we obtain 92.55%, 94.44% and 95.56%, with 10%, 25% and 50% of training pixels.

- The results are obtained with just 49 selected bands for Indian Pines and 37 bands for Salinas, which prove the effectiveness of our algorithm in terms of dimensionality reduction and the selection of relevant bands.

- The results also show the effect of the number of training samples used for classification. We can see that all the metrics (AA, OA and kappa) increases with the size of the training sets for both Indian Pines and Salinas datasets.

- The use of SVM classifier, allowed getting high classification accuracies even with few training samples as in the case of 10% where the OA and kappa achieves respectively 87.61% and 0.8679 for Indian Pines. For Salinas, we get 92.55% in OA and 0.9506 in kappa coefficient.

In the following experiments, the proposed method is compared with two filter band selection algorithms: MRMR and MIH and with a wrapper method based on mutual information. All methods are tested using the same training and testing sets of 50% with SVM-RBF classifier. The obtained results are illustrated in table III for Indian Pines dataset and in table IV for Salinas scene.

The first column in table III and IV represents the total number of samples in each class of the datasets. The remainder columns represent the obtained results of the different methods used in the comparison with the results of our proposed algorithm in the last column. The rows represent the ICA of each class of the scenes. The last rows contains respectively AA, OA and Kappa coefficient.

TABLE III. CLASSIFICATION ACCURACY FOR INDIAN PINES DATASET USING DIFFERENT REPRODUCED METHODS WITH 60 SELECTED BAND

| Class |      | Filter approach | | Wrapper approach | |
|-------|------|------|------|------|------|
|       |      | *MRMR* | *MIH* | *WMIF* | *WNMIPE* |
| 1     | 54   | 47.83 | 82.61 | 73.91 | **82.61** |
| 2     | 1434 | 76.57 | 71.41 | 69.60 | **83.26** |
| 3     | 834  | 60.91 | 78.90 | 66.43 | **83.21** |
| 4     | 234  | 71.79 | 60.68 | 67.52 | **72.65** |
| 5     | 497  | 93.09 | 84.55 | 88.62 | 91.87 |
| 6     | 747  | 96.93 | 90.78 | 92.46 | 95.81 |
| 7     | 26   | 00.00 | 46.15 | 69.23 | **69.23** |
| 8     | 489  | 98.78 | 95.10 | 98.37 | 96.33 |
| 9     | 20   | 00.00 | 60.00 | 70.00 | **80.00** |
| 10    | 968  | 65.08 | 76.65 | 75.62 | **86.36** |
| 11    | 2468 | 89.95 | 81.04 | 87.28 | 87.76 |
| 12    | 614  | 76.22 | 81.11 | 87.30 | 83.39 |
| 13    | 212  | 99.03 | 96.12 | 98.06 | 98.06 |
| 14    | 1294 | 98.30 | 94.74 | 95.83 | 96.29 |
| 15    | 380  | 56.02 | 48.80 | 46.99 | **62.65** |
| 16    | 95   | 91.30 | 93.48 | 93.48 | 91.30 |
| **AA**  |    | 70.11 | 77.63 | 80.04 | **85.05** |
| **OA**  |    | 83.95 | 88.47 | 83.42 | **93.34** |
| **Kappa** |  | 0.8288 | 0.8770 | 0.8231 | **0.9290** |

TABLE IV. CLASSIFICATION ACCURACY FOR SALINAS DATASET USING DIFFERENT REPRODUCED METHODS WITH 37 SELECTED BAND

| Class | | Filter approach | | Wrapper approach | |
|---|---|---|---|---|---|
| | | *MRMR* | *MIH* | *WMIF* | *WNMIPE* |
| 1 | 2009 | 99.20 | 99.50 | 97.71 | **100** |
| 2 | 3726 | 99.95 | 99.89 | 99.78 | **100** |
| 3 | 1976 | 96.36 | 98.89 | 98.99 | **99.90** |
| 4 | 1394 | 99.26 | 99.85 | 99.71 | 99.56 |
| 5 | 2678 | 96.09 | 99.62 | 99.47 | **100** |
| 6 | 3959 | 99.65 | 99.95 | 99.95 | 99.85 |
| 7 | 3579 | 99.33 | 99.72 | 99.61 | **99.83** |
| 8 | **11271** | **89.98** | **85.63** | **96.40** | **91.76** |
| 9 | 6203 | 99.45 | 99.84 | 99.32 | **99.94** |
| 10 | 3278 | 86.27 | 95.36 | 95.91 | **98.72** |
| 11 | **1068** | **86.52** | **98.13** | **97.19** | **100** |
| 12 | 1927 | 99.79 | 99.69 | 99.58 | **100** |
| 13 | 916 | 97.82 | 98.91 | 98.69 | **99.13** |
| 14 | 1070 | 92.90 | 97.20 | 93.08 | **99.44** |
| 15 | 7268 | 44.44 | 60.09 | 53.60 | **78.27** |
| 16 | 1807 | 98.66 | 98.89 | 98.22 | **100** |
| **AA** | | 92.85 | 95.70 | 94.83 | **97.90** |
| **OA** | | 88.71 | 91.40 | 90.31 | **95.56** |
| **Kappa** | | 0.8795 | 0.9083 | 0.8967 | **0.9526** |

- For Indian Pines dataset, it is seen from table III that the proposed method gives better results than the other methods with 85.05% of AA, 93.34% of OA and 0.9290 of Kappa. The proposed WNMIPE exceeds (e.g. in terms of OA) the MRMR with 9.39% the WMIF with 9.92% and the MIH with 4.87%, the latter method combined the mutual information with homogeneity feature for band selection.

- The effectiveness of our proposed method with Indian dataset is more illustrated in figure 3 which shows the ground truth in (a) and the classified maps obtained using the different methods in (b), (c), (d) and (e) for the proposed algorithm. We can see that our method provides the best classified map which confirms the benefit of using the wrapper scheme with NMI and PE to select the relevant bands and eliminate the redundant ones for better classification efficiency.

- For Salinas dataset, from table IV, we can see that the proposed method WNMIPE outperforms the other methods in terms of AA, OA and Kappa. For example, The ICA of the grapes_untrained, which is the class number 8, increases from 85.63% to 91.76% with the proposed algorithm. The ICA of class 11 named lettuce_romaine_4wk increases from 86.52% to 100% which demonstrate the advantage of our method in better finding the optimal set of selected bands to classify the classes of the dataset.

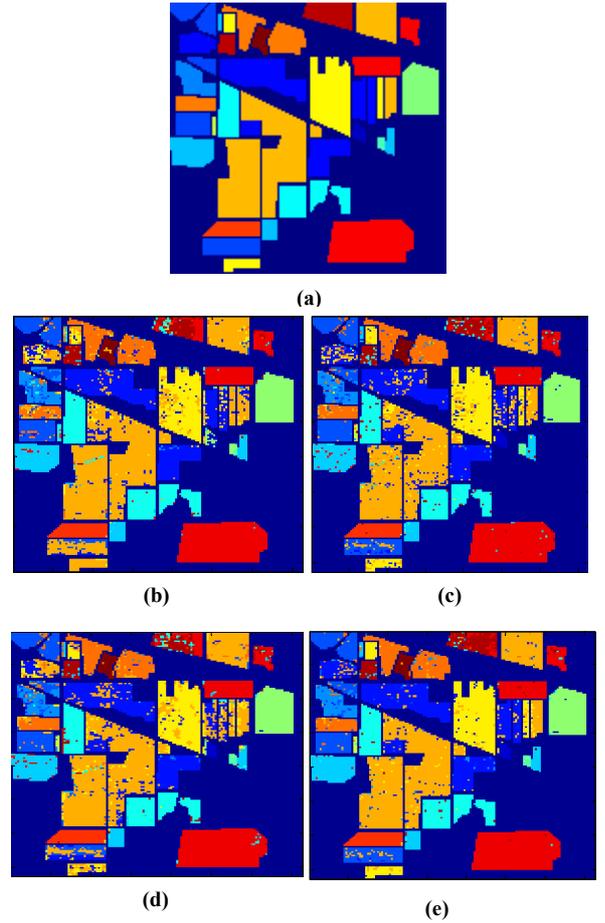

Fig. 3. Indian Pines dataset: (a) Ground truth map; and classified maps with Overall accuracy OA (in parentheses) obtained for 60 selected bands by the reproduced (b) MRMR (83.95), (c) MIH (88.47), (d) WMIF (83.42) and (e) proposed WNMIPE (93.34).

- The classified maps of Salinas dataset using the proposed method and the three reproduced algorithms are shown in Figure 4. It is seen that the best classified map is obtained using our proposed method (e). It is obvious that it is very close to the ground truth shown in sub-figure (a) which proves the efficiency of our algorithm since a reduced set with just 37 bands is sufficient to discriminate all the classes of the scene.

**Partial conclusion**

The analysis of the presented results demonstrates that the proposed algorithm can significantly improve the classification accuracies compared to the other reproduced methods for both Indian Pines and Salinas datasets. Note that the filter methods make the band selection step using the mutual information independently of the classifier system which produces low results than the wrapper methods.

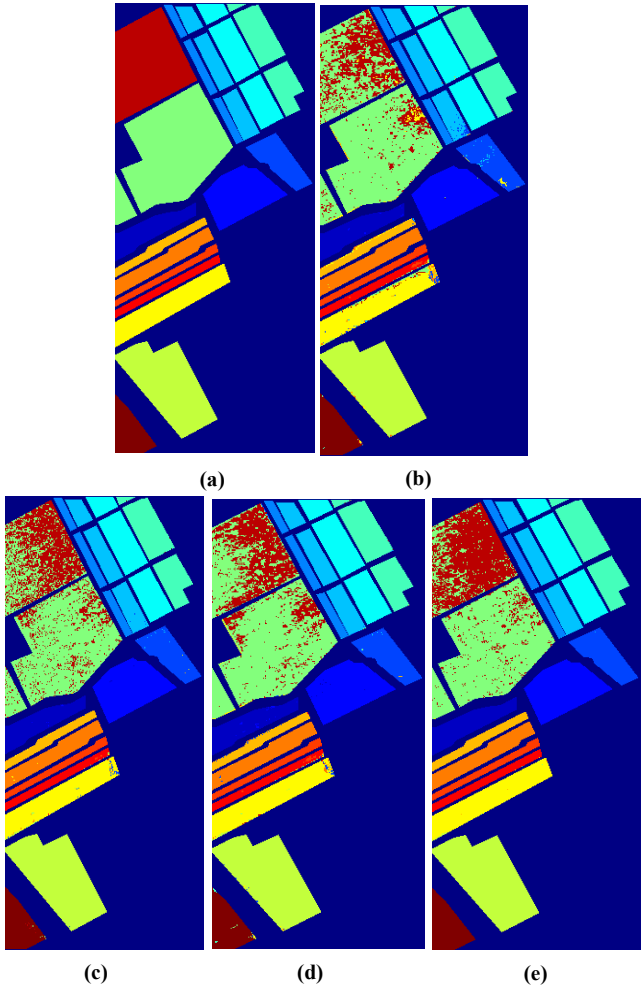

Fig. 4. Salinas dataset: (a) Ground truth map; and classified maps with Overall accuracy OA (in parentheses) obtained for 37 selected bands by the reproduced (b) MRMR (88.71), (c) MIH (91.40), (d) WMIF (90.31) and (e) proposed WNMIPE (95.56).

The MRMR gives the lowest results. The MIH even it is a filter approach, it gives better classification results than WMIF because it incorporates the spatial information presented with the homogeneity feature to select the relevant band but not as well as our proposed method that produced the best results in terms of AA, OA and Kappa coefficient for both Indian Pines and Salinas hyperspectral scenes. The proposed wrapper formed by the combination of the NMI and PE allows selecting relevant bands and eliminating redundant ones for better classes' discrimination and classification efficiency.

## V. Conclusion

The main aim of this paper is to propose a new wrapper approach for dimensionality reduction and land cover classification of hyperspectral images. The algorithm is based on the feature selection using normalized mutual information NMI and error probability PE with SVM classifier. In the selection process, each candidate band had to check two criteria: increases the classification rate and minimizes the probability error to be added in the selected set.

The experiments had been performed on two challenging hyperspectral benchmark datasets with different characteristics captured by the NASA's AVIRIS hyperspectral sensor. The obtained results for tree sets of training samples (10%, 25% and 50%) had been assessed using several classification metrics (ICA, AA, OA and Kappa coefficient), all of them prove that the proposed method can improve the classification accuracies when compared to the state of the art methods and provide accurate classified map.

The classification results using the SVM-RBF classifier achieves 85.05% of AA, 93.34% of OA and 0.9290 of Kappa for Indian dataset. For Salinas scene, we get 97.90% of AA, 95.56% of OA and 0.9526 of Kappa coefficient.

In overall, we can say that the proposed method is an effective tool for HSI classification with reduced number of bands which can be further optimized and improved for example to explore the unlabeled pixels and using unsupervised approaches.

ACKNOWLEDGMENT

The authors would like to thank all the participants involved in this work.